\documentclass[11pt]{article}
\usepackage[preprint]{acl}
\usepackage{times}
\usepackage{latexsym}
\usepackage[T1]{fontenc}
\usepackage[utf8]{inputenc}
\usepackage{inconsolata}
\usepackage{microtype}
\usepackage{booktabs}
\usepackage{amsmath}
\usepackage{algorithm}
\usepackage{algpseudocode}
\usepackage{amssymb}
\usepackage{multirow} 
\usepackage{float}
\usepackage{array}
\usepackage{tikz}
\usepackage{graphicx}
\usepackage{colortbl}
\usepackage{rotating}
\usepackage{adjustbox}
\usepackage{subcaption}
\usepackage{makecell}
\usepackage{tcolorbox}
\tcbuselibrary{most}
\usepackage{pgfplots}
\usetikzlibrary{patterns}
\pgfplotsset{compat=1.18}
\usepackage{placeins}
\usepackage{stfloats}
\usepackage[table]{xcolor}
\definecolor{plotblue}{HTML}{1F77B4}
\definecolor{plotorange}{HTML}{FF7F0E}
\definecolor{plotgreen}{HTML}{2CA02C}
\definecolor{plotred}{HTML}{D62728}
\definecolor{plotpurple}{HTML}{9467BD}
\definecolor{greenDark}{HTML}{1B5E20}   
\definecolor{greenMed}{HTML}{43A047}
\definecolor{greenLight}{HTML}{81C784}
\definecolor{greenPale}{HTML}{E8F5E9} 
\definecolor{refGray}{HTML}{E3E7EC}
\definecolor{greenOne}{HTML}{006D2C} 
\definecolor{greenTwo}{HTML}{159447} 
\definecolor{greenThree}{HTML}{31B85A} 
\definecolor{greenFour}{HTML}{67CF7A} 
\definecolor{greenFive}{HTML}{9BE3A5} 
\definecolor{greenSix}{HTML}{C7F0CB} 
\definecolor{greenSeven}{HTML}{E9F9E9} 
\usepackage{diagbox}

\algnewcommand\algorithmicsetup{\textbf{Hyperparameters:}}
\algnewcommand\Hyperparameters{\item[\algorithmicsetup]}

\newcommand{\name}{\textsc{DORA}~}

\providecommand{\sectionvspace}{\vspace{-0.0cm}}

\title{DORA Explorer: Improving the Exploration Ability of LLMs\\ Without Training}

\author{Priya Gurjar, Md Farhan Ishmam, Kenneth Marino\thanks{Preprint.}\\
Kahlert School of Computing, University of Utah\\
\texttt{\{priya.gurjar,farhan.ishmam,kenneth.marino\}@utah.edu} \\
}

\begin{document}

\maketitle

\begin{abstract}
\vspace{-0.15cm}
Large language model (LLM) agents for sequential decision-making struggle to produce diverse outputs. This leads to insufficient exploration, suboptimal solutions, and repeated actions. Actions are generated at sequence level, but existing sampling strategies, such as temperature scaling, introduce diversity at token level, not at sequence level. We introduce \textsc{DORA Explorer} (Diversity-Oriented Ranking of Actions), a training-free, inference-time algorithm for improving exploration in LLM agents. DORA generates multiple candidate actions, scores them using sequence-level log-probability statistics, and samples an action via a tunable exploration parameter. We first study exploration in the classic Multi-Armed Bandit setting, where DORA substantially outperforms temperature-based sampling. Our main evaluation is on Text Adventure Learning Environment Suite (TALES), where prompting strategies fail to explore but DORA delivers consistent gains across model families, \textit{e.g.}, 31.43\% (ReAct) $\rightarrow$ 45.5\% (DORA) for Qwen-2.5 7B in TextWorld. Beyond exploration, DORA prevents common failures, such as getting stuck in loops.
Our project is available at: 
\url{https://dora-explore.github.io/}.
\end{abstract}
\sectionvspace
\vspace{-0.25cm}
\section{Introduction}
\sectionvspace
The strong generalization capabilities of auto-regressive LLMs have led to their widespread adoption as decision-making agents \cite{wang2024survey}. These tasks require them to act under partial information and balance the acquisition of new knowledge against the exploitation of what they already know. The inability to do so makes LLM agents revisit known sub-optimal actions, settle into local policies, and miss high-reward behavior that requires deliberate exploration~\cite{krishnamurthy2024can, szot2026expanding}.

\begin{figure*}
    \centering
    \includegraphics[width=0.85\linewidth]{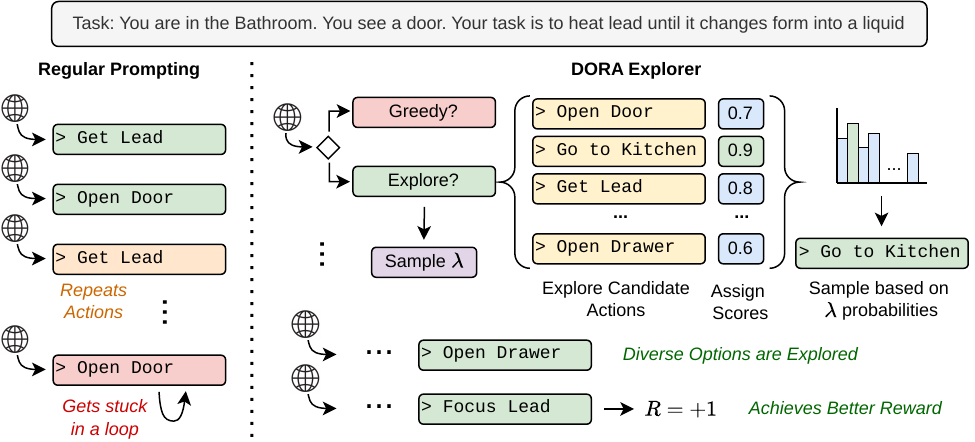}
    \caption{
    (\textit{Left}) Prompting and temperature-based sampling can get stuck in loops by repeating previously explored actions. (\textit{Right}) DORA decides whether to act greedily or explore based on the exploration degree $\lambda$, generates candidate actions, scores them, and samples via the derived $\lambda$-probabilities, resulting in better terminal rewards.}
    \vspace{-0.25cm}
    \label{fig:teaser}
\end{figure*}

In environments with text interfaces, LLM agents use natural language as the output space to produce domain-specific actions. While this allows agents to operate zero-shot~\cite{huang2022language}, it means that proper exploration requires sampling diverse \emph{sequences of tokens}, not just diverse individual tokens. The sequences "\texttt{open the wooden door}" and "\texttt{open the wooden box}" differ by one token but lead to different trajectories.

Sampling diverse sequences is challenging and has been well explored in the narrative generation literature~\cite{peeperkorn2024temperature}. LLMs continue to generate high-likelihood and low-diversity outputs, resulting in repetitive action patterns~\cite{krishnamurthy2024can}. Temperature scaling is the most common strategy for increasing diversity, but \citet{peeperkorn2024temperature} found only a weak correlation between temperature and novelty in narrative generation. \citet{shi2024thorough} showed that raising temperature degrades task accuracy.

We introduce \textsc{DORA Explorer} (Diversity-Oriented Ranking of Actions), a training-free, inference-time framework for improving exploration in LLM agents (Figure~\ref{fig:teaser}). 
At each step, DORA generates a pool of candidate actions, scores them using sequence-level log-probability statistics, and samples according to an exploration degree $\lambda$.
DORA's primary variant, \emph{Auto-Explore}, lets the agent autonomously decide at each step whether to act greedily or explore, and at what intensity. We use the Text Adventure Learning Environment Suite (TALES)~\cite{cui2025tales} as our main evaluation benchmark, in which DORA Auto-Explore consistently outperforms sampling baselines across model families and sizes. We also study a scheduled variant of DORA in the classic Multi-Armed Bandit (MAB) problem to isolate these exploration dynamics in a simpler setting. DORA produces strong gains over greedy and sampling baselines across all metrics.

We summarize the contributions made in our work. First, we provide an analysis showing that current decoding strategies (temperature scaling) and prompting-based methods fail to produce robust, sequence-level exploration, in both the classic MAB setting and the more realistic TALES benchmark (\S\ref{sec:experiments}). Second, we introduce DORA Explorer, a training-free, inference-time algorithm that generates diverse candidate actions, scores them using sequence-level log-probability statistics, and samples using a tunable exploration parameter $\lambda$. Its primary variant, Auto-Explore, requires no per-environment configuration (\S\ref{sec:methodology}). Third, we show that DORA delivers consistent gains across TALES environments and model families (\S\ref{sec:tales_results}). We conduct ablations that isolate the contributions of DORA's scoring components and exploration-selection mechanism, showing its gains stem from structured exploration rather than generic stochasticity (\S\ref{sec:ablations}).

\sectionvspace
\section{Related Work}
\sectionvspace
We categorize four approaches to controlling LLM agent exploration: sampling-based/controlled decoding, prompting-based methods, inference-time frameworks, and training-based approaches. DORA draws on the first three, providing a framework that explicitly controls exploration and samples diverse actions without additional training.

\paragraph{Sampling and controlled decoding.} Decoding strategies such as top-$k$~\cite{fan2018hierarchical}, top-$p$~\cite{holtzman2019curious}, and min-$p$~\cite{nguyen2024turning} sampling target fluency and repetition rather than exploration: temperature has little effect on accuracy below $\tau=1$~\cite{renze2024effect} and only weakly correlates with novelty~\cite{peeperkorn2024temperature}. \citet{troshin2025control} switch between greedy and high-temperature token-level sampling via a learned, task-specific classifier, demonstrated only on verifiable math tasks. DORA instead controls exploration at the sequence level, decoupling candidate generation from $\lambda$-based sampling and requiring no training or task-specific classifier.

\paragraph{Prompting-based methods.} \citet{krishnamurthy2024can} find that only one of 32 prompt designs: GPT-4 with CoT over an externally summarized history, achieves robust MAB exploration, suggesting prompting-only solutions are fragile and hard to generalize. Verbalized Sampling~\cite{zhang2025verbalized} generates diverse outputs with associated probabilities; DORA uses a similar technique for candidate generation (\S\ref{subsec:actionGeneration}). CoT~\cite{wei2022chain} and ToT~\cite{yao2023tree} improve reasoning but do not target exploration under uncertainty.

\paragraph{Inference-time exploration frameworks.} Without modifying model weights, EAST~\cite{rahn2024controlling} steers exploration via a precomputed activation vector, \citet{arumugam2025toward} implements Posterior Sampling for RL through orchestrated LLM subroutines, and OLIVIA~\cite{yu2026olivia} places a UCB-based contextual bandit at the action-selection layer of a frozen ReAct agent. DORA requires none of these auxiliary structures, operating purely on the base model's own candidate actions and log-probabilities.

\paragraph{Training-based approaches.} EVOLvE~\cite{nie2024evolve} distills classical exploration algorithms into LLMs via in-context and fine-tuning methods, with fine-tuning yielding the strongest gains, \citet{schmied2025llms} use RL on self-generated CoT rationales to correct greedy action selection, \citet{tajwar2025training} train models to explore and adapt directly. These methods require model-specific training; DORA is inference-time and complementary and could, in principle, be applied atop a fine-tuned policy.
\sectionvspace
\section{Methodology}
\label{sec:methodology}
\sectionvspace
\subsection{Problem Formulation}
\vspace{-0.1cm}
\label{sec:problemFormulation}
We model the multi-step decision-making task as an episodic Partially Observable Markov Decision Process (POMDP), defined by the 7-tuple $\langle\mathcal{S},\mathcal{A},\mathcal{O}, \mathbb{T}, \mathbb{T}_0, \mathbb{O}, R\rangle$. At step $t$, the agent receives the observation $o_t\in\mathcal{O}$, where $o_t\sim\mathbb{O}(\cdot\mid s_t)$, for latent state $s_t\in\mathcal{S}$, selects an action $a_t\in\mathcal{A}$, and receives reward $r_t = R(s_t,a_t)$. The environment then transitions as $s_{t+1}\sim\mathbb{T}(\cdot\mid s_t,a_t)$, with $\mathbb{T}_0$ the initial state distribution. Since $s_t$ is unobserved, the agent instead conditions on the interaction history $h_t=\{(o_i,a_i,r_i)\}_{i=1}^{t-1}$, acting under the history-conditioned autoregressive policy $\pi(a\mid h_t, \tau;\theta)$ with parameters $\theta$ and temperature $\tau$. An LLM serves as this policy. Given $h_t$ and $o_t$, it autoregressively generates a token sequence $y_t$, which is parsed into an executable action $a_t = \psi(y_t) \in \mathcal{A}_t$. Unparsable or infeasible generations are marked invalid.

\subsection{Exploration via Temperature Scaling}
\label{sec:tempScalingMethodology}
For autoregressive agents, diversity and creativity in generated outputs are commonly controlled via temperature scaling, which rescales the model's raw output scores (logits) prior to the softmax operation~\citep{troshin2025control}. 
Temperature scaling is often loosely described as a form of exploration, echoing the reinforcement learning notion of an agent visiting a broader range of states. However, temperature scaling operates purely at the \emph{token} level: raising $\tau$ increases the entropy of each per-token distribution, but a sequence sampled from a higher-entropy distribution need not correspond to a materially different \emph{action}. We make this concrete below.
Given the context $c = (x_1,\dots,x_{|c|})$, the model autoregressively generates a continuation $\mathbf{x} = (x_{|c|+1}, \dots, x_{|c|+L})$ of length $L$, with joint probability
\vspace{-0.25cm}
\begin{equation}
P(\mathbf{x} \mid c, \tau) = \prod_{l=1}^{L} P_\tau(x_{|c|+l} \mid c, x_{<|c|+l}),
\vspace{-0.15cm}
\end{equation}
where $j = |c|+l$ is the target position. Each conditional is given by the temperature-scaled softmax over the vocabulary,
\vspace{-0.25cm}
\begin{equation}
\label{eq:softmaxProbabilities}
P_\tau(x_j = v \mid c, x_{<j}) = \frac{\exp\!\left(z_v^{j} / \tau\right)}{\sum_{k=1}^{V} \exp\!\left(z_k^{j} / \tau\right)},
\vspace{-0.25cm}
\end{equation}
for $v \in \{1,\dots,V\}$, where $z_i^{j}$ denotes the $i^\text{th}$ logit at position $j$ and $\tau$ controls the sharpness of the resulting distribution: at low $\tau$, the model greedily selects its most probable continuation, while at high $\tau$, the distribution approaches uniform and the model samples increasingly unlikely tokens. In practice, temperature scaling is rarely used alone, since sampling from an unconstrained high-entropy distribution risks selecting low-quality tokens; LLMs therefore typically pair it with a token-filtering strategy such as top-$p$ (nucleus) sampling~\citep{Holtzman2020The}.

Critically, none of this machinery controls diversity at the level the agent actually acts on. A sequence sampled at high $\tau$ may differ in wording while still parsing into the same action, or may simply be invalid; either way, token-level entropy does not guarantee the agent visits a different or more informative state. Robust exploration instead requires diversity among complete candidate \emph{actions}.

\sectionvspace
\subsection{DORA Explorer}
\vspace{0.10cm}
\sectionvspace
DORA Explorer realizes exploration through three composed mechanisms: (i) diverse, deduplicated \emph{candidate generation} (\S\ref{subsec:actionGeneration}), which guarantees novelty before any scoring; (ii) a \emph{scoring function} (\S\ref{subsec:scoring}) combining a likelihood-based confidence estimate with a within-sequence coherence signal, and (iii) \emph{$\lambda$-controlled Boltzmann sampling}, which governs how sharply candidates are exploited. The degree of exploration $\lambda$ can be set by a fixed schedule or by the agent itself (\S\ref{subsec:selectLambda}). Algorithm~\ref{alg:dora} formalizes the method, with prompts in \S\ref{app:prompts}.

\sectionvspace
\vspace{0.10cm}
\begin{algorithm}
\caption{Diversity-Oriented Ranking of Action (DORA) Explore Algorithm}
\label{alg:dora}
\small
\begin{algorithmic}[1]
\setlength{\itemsep}{0pt}
\setlength{\parsep}{0pt}
\Require History $\mathcal{H}$, current observation $o_{t}$, previously used action set $\mathcal{A}_{o_t}$
\Hyperparameters Candidate count $n_\mathcal{C}$, temperatures $\{\tau_d, \tau_\lambda, \tau_\mathcal{C}\}$, diversity weight $\alpha$
\Ensure Selected action $a^*_t$ at step $t$
\State $c\leftarrow [\mathcal{H};o_t]$ \Comment{Concatenate current observation}
\State $d \sim \pi_d(c,\; \tau_d)$
\Comment{Decide greedy or explore}
\If{$d = \textsc{Explore}$}
    \State $\lambda \leftarrow \begin{cases} \lambda_{\text{exp}}(t) & \text{if using scheduler} \\ \pi_{\lambda}(c, \tau_\lambda) & \text{if using policy} \end{cases}$
    \State $\mathcal{C} \leftarrow \emptyset$ 
    \Comment{Initialize candidate action set}
    \State $\mathcal{C}' \sim \pi_\mathcal{C}(c,\tau_\mathcal{C}, n_\mathcal{C})$  \Comment{Sample $n_\mathcal{C}$ possible actions}
    \For{$a$ \textbf{in} $\mathcal{C}'$}
        \If{$a\notin\mathcal{A}_{o_t}$} \Comment{New action} 
        \State $\mathcal{C} \leftarrow \mathcal{C} \cup \{a\}$ 
        \EndIf
    \EndFor
\State $\mathbf{s} = (\text{Score}(a, \alpha))_{a\in\mathcal{C}}$ \Comment{Generate the score vector}
\State $\mathbf{p}_\lambda = \text{Softmax}(\lambda\cdot\mathbf{s})$ \Comment{Generate $\lambda$ probabilities}
\State $a^*_t \sim \text{Categorical}(\mathbf{p}_\lambda)$
    \Comment{Sample based on $\lambda$ prob.}
\Else
    \State $a^*_t \sim \pi_g(c,\; \tau{=}0)$
\Comment{Sample greedy action}
\EndIf
\State $\mathcal{A}_{o_t}\leftarrow\mathcal{A}_{o_t}\cup\{a^*_t\}$
\State \Return $a^*_t$
\end{algorithmic}
\end{algorithm}
\subsubsection{Candidate Action Generation}
\label{subsec:actionGeneration}
At each step $t$, given context $c = [\mathcal{H}; o_t]$, the agent samples a greedy-exploration decision $d \sim \pi_d(c, \tau_d)$. If exploring, it samples $n_\mathcal{C}$ candidate actions jointly via list-level prompting $\pi_\mathcal{C}(c,\tau_\mathcal{C},n_\mathcal{C})$. Repeated single-action sampling collapses to the same high-probability mode. Prompting for a list of $n_\mathcal{C}$ actions \emph{jointly} instead forces coverage of the output distribution's breadth, naturally yielding diverse candidates~\citep{zhang2025verbalized}. Candidates already used at the current observation are discarded, so action novelty is enforced before scoring (\S\ref{subsec:scoring}). If the agent is not exploring, the action is sampled greedily, $a^*_t\sim\pi_g(c,\tau{=}0)$.
\sectionvspace
\vspace{0.10cm}
\subsubsection{Action Sequence Scoring and Sampling}
\label{subsec:scoring}
Each candidate $a\in\mathcal{C}$ is a token sequence $a=\langle x_1,\dots,x_n\rangle$, scored as
\begin{equation}
\label{eq:score}
    s = \text{Score}(a,\alpha) = \alpha\,\tilde{\mu}(a) - (1-\alpha)\,\tilde{\sigma}^2(a),
\end{equation}
where $\alpha$ is a scoring hyperparameter and $\tilde{\mu}, \tilde{\sigma}^2$ are the min-max normalized mean and variance of per-token log-probabilities (Eq.~\ref{eq:softmaxProbabilities}) (\textit{c.f.}, \S\ref{subsec:minMaxNormalization}).
Given scores $\mathbf{s}=(s_a)_{a\in\mathcal{C}}$, we sample via
\begin{equation}
\label{eq:lambdaProbabilities}
    P_{\lambda}(a) = \text{Softmax}(\lambda \, s_a) = \frac{\exp(\lambda \, s_a)}{\sum_{a'\in\mathcal{C}}\exp(\lambda \, s_{a'})},
\end{equation}
with $a^*_t \sim \text{Categorical}(\mathbf{p}_\lambda)$. This is Boltzmann (Gibbs) action selection, $\pi(a)\propto\exp(Q(a)/\tau)$, a standard RL exploration strategy~\citep{sutton2018reinforcement}, with $s_a$ as $Q(a)$ and $\lambda$ the inverse temperature. Lower $\lambda$ flattens $P_\lambda$ toward uniform sampling (exploration), while higher $\lambda$ concentrates mass on the top candidate (exploitation). Unlike token-level temperature (Eq.~\ref{eq:softmaxProbabilities}), which reshapes the distribution over the {next token}, $\lambda$ reshapes the distribution over \emph{complete} and \emph{diverse} action sequences.

\subsection{Why Mean and Variance of Log-Probabilities?}

\sectionvspace

As DORA is training-free and lacks access to a learned $Q$-value, we use the policy's length-normalized log-probability as a practical signal of action quality. This choice is consistent with maximum-entropy control, where an action's likelihood under the original policy contributes to its utility~\citep{levine2018reinforcement,haarnoja2017reinforcement,haarnoja2018soft}. Although the mean log-probability $\tilde{\mu}(a)$ is not an exact measure of epistemic uncertainty, it provides a tractable confidence signal commonly used for uncertainty estimation, correctness prediction, and sequence scoring~\citep{malinin2021uncertainty,mahaut2024factual}. The token-level variance $\tilde{\sigma}^2(a)$ provides complementary information about a candidate's internal consistency: high variance indicates that some tokens are much less supported than others; we observe that such sequences are often malformed or incoherent. DORA scoring thus favors candidates with high average likelihood and low token-level variance. Because candidate diversity is already encouraged during action generation (\S\ref{subsec:actionGeneration}), the variance penalty primarily filters unreliable actions rather than suppressing exploration.

\sectionvspace
\subsection{How do we select $\lambda$?}
\label{subsec:selectLambda}
\vspace{-0.15cm}
The fixed schedule and agent-driven policy below are two instantiations of the same $\lambda$-control mechanism. Where the exploration$\to$exploitation shift is predictable, a schedule suffices. We use
\begin{equation}
\label{eq:lambdaExpSchedule}
    \lambda_\text{exp}(t) = \lambda_{\min} + (\lambda_{\max}-\lambda_{\min})\frac{e^{kt/T}-1}{e^k-1},
\end{equation}
with horizon $T$ and growth constant $k$. While this suits MAB's predictable structure, it is impractical when exploration fluctuates unpredictably, \textit{e.g.}, in TALES (\S\ref{sec:tales_results}). We instead let the agent sample $\lambda \sim \pi_\lambda(c, \tau_\lambda)$ at each step, conditioned on the context $c$. This \emph{Auto-Explore} variant uses a single fixed hyperparameter configuration (Table~\ref{tab:tales_hyperparams}) across all TALES environments and models, with no environment-specific tuning. Auto-Explore is our primary, recommended configuration, and the scheduled variant is a lighter-weight alternative for predictable settings like MAB (\S\ref{sec:mab_results}).
\sectionvspace
\vspace{0.10cm}
\section{Experiments}
\label{sec:experiments}
\sectionvspace
\vspace{0.10cm}
\subsection{Multi-Armed Bandits (MAB)}
\label{sec:mab_results}
\sectionvspace
\vspace{0.20cm}
\paragraph{Experimental Setup.} Multi-Armed Bandits (MAB) provide a classical testbed for studying the exploration-exploitation trade-off~\citep{mehlhorn2015unpacking,sutton1998reinforcement}. As a single-state Markov Decision Process, MAB serves as a minimal ``unit test'' for evaluating new exploration strategies in decision-making. Following \citet{krishnamurthy2024can}, we consider stochastic Bernoulli bandits with $K$ arms, where each arm $a \in [K]$ has an unknown mean reward $\mu_a \in [0,1]$. At each time step $t \in [T]$, the agent selects an arm $a_t$ and observes a reward $r_t \sim \mathrm{Bernoulli}(\mu_{a_t})$. The objective is to maximize cumulative reward, or equivalently, minimize cumulative regret,
\vspace{-0.15cm}
\begin{equation}
R_T = T \mu^\star - \mathbb{E}\!\left[\sum_{t=1}^{T} \mu_{a_t}\right].
\vspace{-0.25cm}
\end{equation}
\textbf{Baselines.} We evaluate MAB using \texttt{Llama-3.1 8B}, results reported in \autoref{tab:mab}, and \texttt{Qwen-2.5 7B}, results in Appendix~\ref{app:mab_results}. We compare against classical bandit algorithms, UCB~\citep{auer2002finite} and Thompson Sampling (TS)~\citep{agrawal2012analysis}, as well as Greedy and $\epsilon$-greedy strategies. Unlike LLM-based methods, which must select actions through language and are therefore subject to sampling variance, invalid or malformed outputs, and parsing text into discrete actions, these classical algorithms act directly on the action space and require no such mediation. UCB and TS further enjoy theoretical performance guarantees, making them natural upper-bound references rather than baselines. To contextualize LLM behavior, we include sampling-based policies with fixed temperatures $(0, 0.7, 1.0, 1.5)$ and an adaptive $\tau$-policy with exponentially decaying temperature, mirroring the $\lambda$-schedule (Eq.~\ref{eq:lambdaExpSchedule}). 

To adapt DORA to MAB, we use a scheduled $\lambda$ policy (cf.~\S\ref{subsec:selectLambda}), exponentially increasing $\lambda \in [0, 40]$ to transition from early exploration to later exploitation. We report mean average reward, cumulative regret, fraction of optimal arm selections, and \textit{Suffix Failure Frequency} --- the fraction of runs failing to select the optimal arm at least once after $T/2$ steps.

\sectionvspace
\subsubsection{Results and Analysis}
\sectionvspace
\begin{table*}[t] \vspace{-0.1cm} \centering \setlength{\tabcolsep}{4pt} \renewcommand{\arraystretch}{1.6} \begin{adjustbox}{max width=\textwidth} \begin{tabular}{l*{2}{c}|*{7}{c}c} \toprule \multirow{2}{*}{ \diagbox[width=8em,height=3em] {\textbf{Metric}}{\textbf{Method}} } & \multicolumn{2}{c|}{\textit{Reference (Upper Bound)}} & \multicolumn{7}{c}{Baselines} & \multirow{2}{*}{ \begin{tabular}{c} \textbf{DORA}\\ \textbf{$\lambda_\text{exp}$} \end{tabular} }\\ \cmidrule(lr){2-3} \cmidrule(lr){4-10} & UCB & TS & Greedy & $\varepsilon$-Greedy & $\tau=0$ & $\tau=0.7$ & $\tau=1$ & $\tau=1.5$ & $\tau_\text{exp}$ & \\ \midrule Mean Avg Reward & \cellcolor{refGray}0.564 & \cellcolor{refGray}0.548 & \cellcolor{greenFour}\textbf{0.503} & \cellcolor{greenThree}\textbf{\textcolor{white}{0.509}} & \cellcolor{greenSeven}\textbf{0.414} & \cellcolor{greenSeven}\textbf{0.414} & \cellcolor{greenSix}\textbf{0.422} & \cellcolor{greenTwo}\textbf{\textcolor{white}{0.518}} & \cellcolor{greenFive}\textbf{0.484} & \cellcolor{greenOne}\textbf{\textcolor{white}{0.538}} \\ SuffFailFreq($T$/2) & \cellcolor{refGray}0.011 & \cellcolor{refGray}0.000 & \cellcolor{greenThree}\textbf{\textcolor{white}{0.465}} & \cellcolor{greenTwo}\textbf{\textcolor{white}{0.230}} & \cellcolor{greenFive}\textbf{0.900} & \cellcolor{greenFive}\textbf{0.900} & \cellcolor{greenFour}\textbf{0.850} & \cellcolor{greenOne}\textbf{\textcolor{white}{0.000}} & \cellcolor{greenOne}\textbf{\textcolor{white}{0.000}} & \cellcolor{greenOne}\textbf{\textcolor{white}{0.000}} \\ Best arm Frac & \cellcolor{refGray}0.824 & \cellcolor{refGray}0.743 & \cellcolor{greenFive}\textbf{0.515} & \cellcolor{greenFour}\textbf{0.550} & \cellcolor{greenSeven}\textbf{0.100} & \cellcolor{greenSeven}\textbf{0.100} & \cellcolor{greenSix}\textbf{0.131} & \cellcolor{greenThree}\textbf{\textcolor{white}{0.675}} & \cellcolor{greenTwo}\textbf{\textcolor{white}{0.698}} & \cellcolor{greenOne}\textbf{\textcolor{white}{0.703}} \\ Cum Regret & \cellcolor{refGray}17.637 & \cellcolor{refGray}25.686 & \cellcolor{greenFive}\textbf{48.502} & \cellcolor{greenFour}\textbf{44.998} & \cellcolor{greenSeven}\textbf{90.000} & \cellcolor{greenSeven}\textbf{90.000} & \cellcolor{greenSix}\textbf{86.900} & \cellcolor{greenTwo}\textbf{\textcolor{white}{34.640}} & \cellcolor{greenThree}\textbf{\textcolor{white}{37.160}} & \cellcolor{greenOne}\textbf{\textcolor{white}{28.730}} \\ \bottomrule \end{tabular} \end{adjustbox} \vspace{-0.25cm} 
\caption{ Performance on the Bernoulli MAB instance using \texttt{Llama-3.1 8B} with horizon $T=500$. Grey columns denote reference algorithms; darker green indicates better performance among LLM-based methods. UCB, TS, Greedy, $\epsilon$-greedy are averaged over N = 1000 runs, while
LLM-based methods are averaged over N = 20
runs.} 
\vspace{-0.55cm} 
\label{tab:mab} 
\end{table*}

\textbf{Failure Modes of Temperature Scaling.} As shown in \autoref{tab:mab}, fixed temperature policies ($\tau \in \{0, 0.7, 1, 1.5\}$) exhibit two failure modes. For $\tau \leq 1$, the model behaves overly greedily, committing early to a suboptimal arm; this yields high \textit{Suffix Failure Frequency} (up to 95\%). For $\tau > 1$, increased randomness reduces suffix failure but prevents consistent exploitation and increases invalid actions. An adaptive $\tau$-policy decaying from $\tau=2$ to $\tau=0$ reduces suffix failures and improves regret relative to fixed temperatures, but still fails to reliably exploit the optimal arm.\\

\begin{figure}[t]
\centering
\includegraphics[width=\columnwidth]{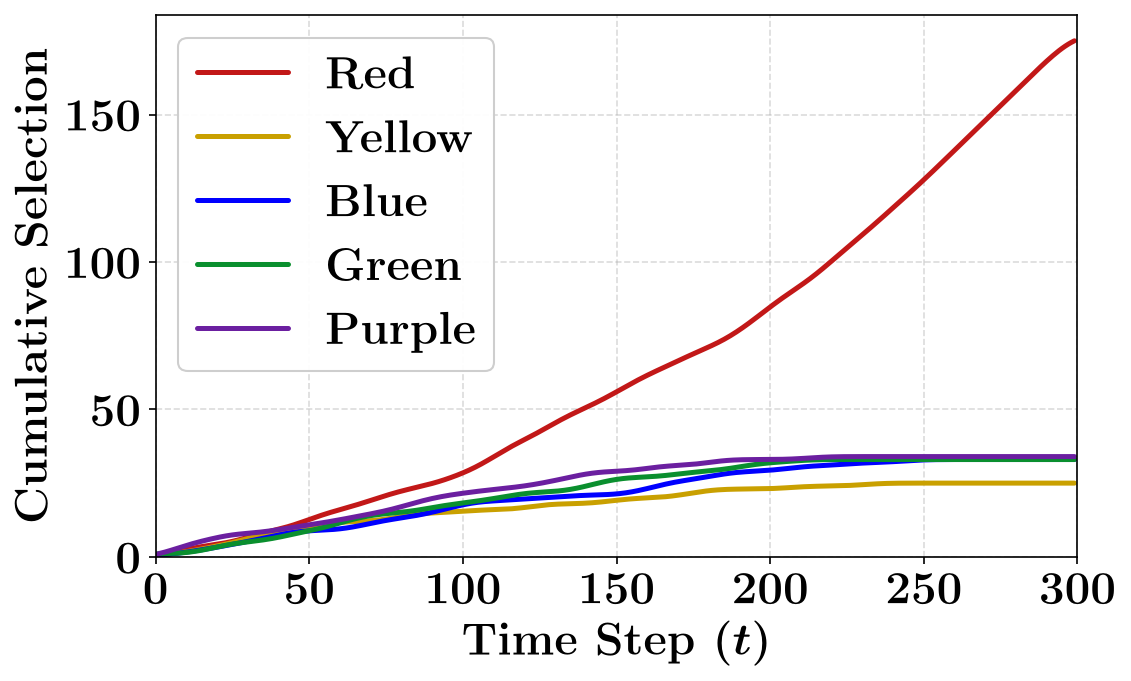}
\caption{
Arms selected by DORA over an episode in Llama 3.1 8B. Early in the episode, the
$\lambda$-schedule (\S\ref{subsec:selectLambda}) encourages broad exploration across candidate actions. As $\lambda$ increases, DORA progressively shifts toward exploitation, selecting the highest-scoring arm (red) more frequently.
}
\label{fig:lambda_schedule}
\vspace{-0.3cm}
\end{figure}

\noindent\textbf{Scheduled $\lambda$ Enables Controlled Exploration.} Our scheduled $\lambda$ policy (cf.~\S\ref{subsec:selectLambda}) provides fine-grained control over the exploration--exploitation trade-off. As shown in \autoref{fig:lambda_schedule}, lower $\lambda$ induces exploratory behavior early on, sampling across arms, while higher $\lambda$ later concentrates probability mass on high-confidence actions.

\begin{figure}[t]
\centering
\includegraphics[width=\columnwidth]
{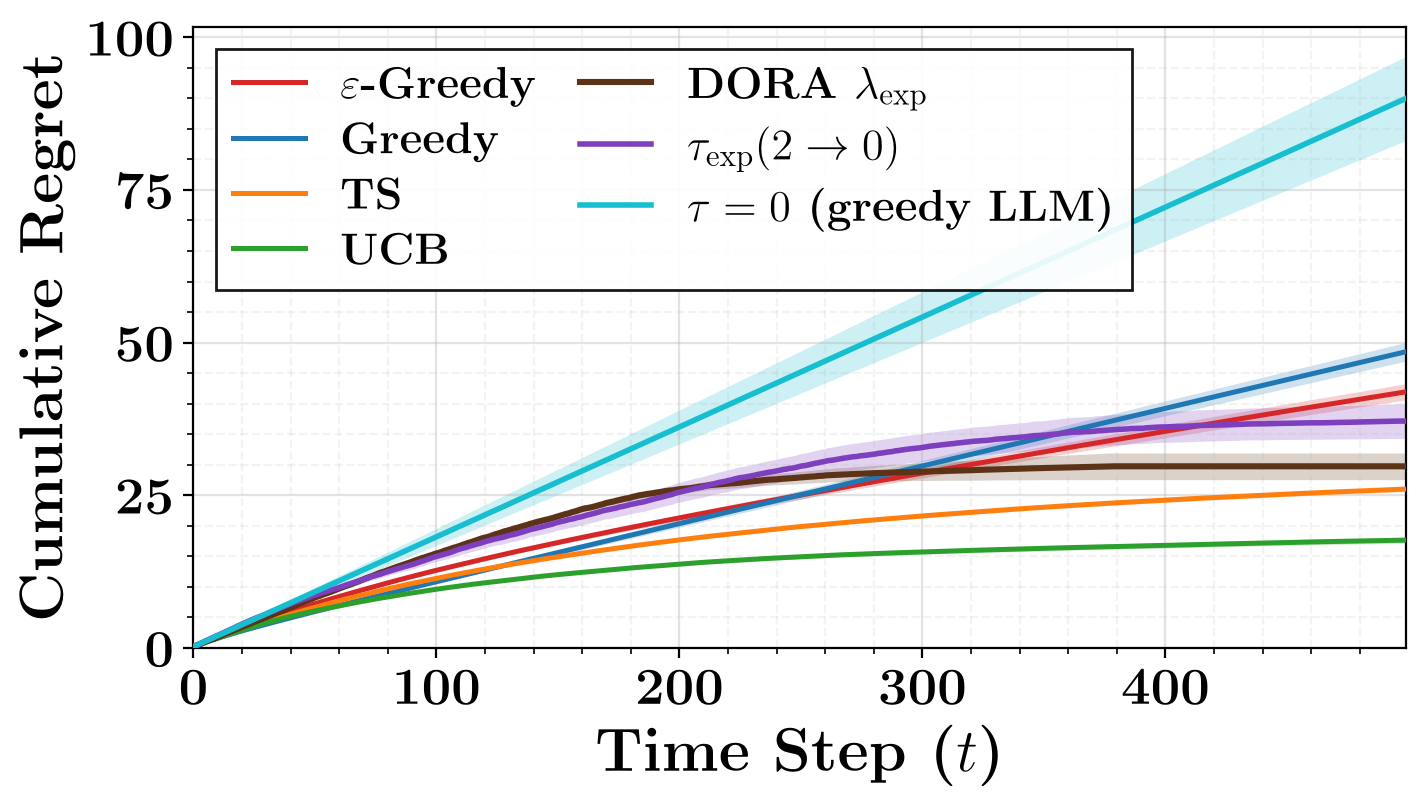}
\vspace{-0.3cm}
\caption{
Cumulative regret over a horizon of $T=500$ for Llama 3.1 8B.
DORA substantially reduces regret relative to all LLM-based sampling baselines.
}
\label{fig:regret_curve}
\vspace{-0.45cm}
\end{figure}
\paragraph{DORA Closes the Gap to Classical Exploration Algorithms.} As shown in \autoref{fig:regret_curve} and \autoref{tab:mab}, DORA achieves substantially lower cumulative regret than all LLM-based sampling baselines, zero \textit{Suffix Failure Frequency}, approaching the performance of UCB and Thompson Sampling, both theoretical upper bounds, without any bandit-specific statistics. This trend holds true across different time horizons, additional arm counts, and reward distributions (\S\ref{app:mab_results}).

\subsection{TALES}
\label{sec:tales_results}
\sectionvspace
TALES~\citep{cui2025tales} is a unified benchmark spanning five text-based environments, TextWorld, TextWorldExpress, AlfWorld, ScienceWorld, and Jericho, that vary in horizon length, parser strictness, feedback quality, and reward structure. Because TALES preserves each environment in its canonical form with minimal task-specific scaffolding, it provides a direct test of an agent's ability to reason, plan, and explore from text alone under uncertainty. Each TALES task is a POMDP (\S\ref{sec:problemFormulation}), where the agent receives a textual observation along with interaction history and outputs a textual action at each step. Details in \S~\ref{app:talessetup}.\\\vspace{-0.2cm}

\noindent\textbf{Environments.} TextWorld uses CookingWorld instances~\citep{cote2018textworld}, in which the agent follows recipes by locating and processing ingredients as task complexity increases. TextWorldExpress~\citep{jansen2023textworldexpress} is a faster re-implementation with a stricter parser and simpler deterministic tasks. AlfWorld~\citep{shridhar2020alfworld} consists of text-based household tasks with sparse feedback, no intermediate rewards, and the constraint that the agent can hold only one object at a time. ScienceWorld~\citep{wang2022scienceworld} contains open-ended scientific tasks with multiple valid solution paths and occasional irreversible states. Jericho~\citep{hausknecht2020interactive} includes long-horizon, human-authored interactive fiction and is the most challenging setting in the suite.\\

\noindent\textbf{Experimental Setup.} Given the dynamic exploration requirements of TALES tasks, we use the \emph{DORA Auto-Explorer} variant, as described in Algorithm~\ref{alg:dora}. The LLM dynamically decides whether to explore and, if so, selects an exploration parameter $\lambda$ at each step, enabling intermittent exploration behavior within an episode.\\

\noindent\textbf{Baselines and Metrics.} We compare DORA against zero-shot, chain-of-thought (CoT)~\citep{wei2022chain}, tree-of-thought (ToT)~\citep{yao2023tree}, ReAct~\citep{yao2022react}, and an exploration-based prompt (\S\ref{app:tales_prompt}). Since raw scores vary across games, we normalize each score by the maximum achievable score for the corresponding task and report mean percentage scores within each environment. Results are averaged over three random seeds and reported as mean $\pm$ standard error.

\begin{figure}[t]
\centering
\begin{tikzpicture}
\begin{axis}[
    ybar,
    width=\columnwidth,
    height=5cm,
    enlarge x limits={abs=0.85cm},
    ylabel={\#Unique States},
    ymin=0,
    ymax=35,
    grid=both,
    major grid style={dashed, gray!55},
    minor grid style={gray!20},
    minor tick num=1,
    symbolic x coords={TextWorld, ScienceWorld, AlfWorld},
    xtick=data,
    xticklabel style={rotate=25, anchor=north east},
    legend image code/.code={
        \draw[#1, draw=black] (0cm,-0.12cm) rectangle (0.15cm,0.12cm);
    },
    tick label style={font=\small},
    bar width=5.5pt,
    legend style={
        at={(0.5,1.05)},
        anchor=south,
        legend columns=3,
        font=\footnotesize,
        draw=none
    },
]

\addplot[fill=cyan!17, draw=black!80] coordinates {
(TextWorld,14.0) (ScienceWorld,6.5) (AlfWorld,2.67)
};
\addplot[fill=orange!15, draw=black!80] coordinates {
(TextWorld,12.1) (ScienceWorld,8.9) (AlfWorld,3.67)
};
\addplot[fill=teal!15, draw=black!80] coordinates {
(TextWorld,17.5) (ScienceWorld,8.23) (AlfWorld,3.25)
};
\addplot[fill=blue!15, draw=black!80] coordinates {
(TextWorld,12.1) (ScienceWorld,4.6) (AlfWorld,3.67)
};
\addplot[fill=red!15, draw=black!80] coordinates {
(TextWorld,22.0) (ScienceWorld,13.77) (AlfWorld,3.33)
};
\addplot[
fill=magenta!17,
draw=black!80,
postaction={
pattern=north east lines,
pattern color=black!70
}
] coordinates {
(TextWorld,32.0) (ScienceWorld,20.23) (AlfWorld,27.92)
};

\legend{Zero-shot, Prompt Explore, CoT, ToT, ReAct, DORA}

\end{axis}
\end{tikzpicture}

\vspace{-0.2cm}
\caption{
Average unique states visited per task in TALES environments for LLama 3.1 8B. DORA explores substantially more of the environment than prompting methods.
}
\label{fig:state_diversity}
\vspace{-0.35cm}
\end{figure}

\sectionvspace
\subsubsection{TALES Results and Analysis}
\begin{table*}[ht]
    \centering
    \small
    \setlength{\tabcolsep}{3pt}
    \begin{tabular}{llccccc}
        \toprule
        \textbf{Model} & \textbf{Setting} & \textbf{TextWorld} & \textbf{TW Express} & \textbf{AlfWorld} & \textbf{ScienceWorld} & \textbf{Jericho} \\
        \midrule
        \multirow{6}{*}{Qwen-2.5 7B} 
            & Zero-shot & $29.20_{\pm 1.92}$ & $48.93_{\pm 1.69}$ & $0.00_{\pm 0.00}$ & $13.49_{\pm 0.12}$ & $0.66_{\pm 0.04}$ \\
            & Chain of Thought & $24.89_{\pm 0.91}$ & $37.44_{\pm 0.00}$ & $0.00_{\pm 0.00}$ & $09.81_{\pm 0.18}$ & $1.58_{\pm 0.03}$ \\
            & Tree of Thought & $11.39_{\pm 0.83}$ & $45.90_{\pm 0.21}$ & $0.00_{\pm 0.00}$ & $10.10_{\pm 0.00}$ & $1.40_{\pm 0.00}$\\
            & Prompt Explore & $23.12_{\pm 1.29}$ & $47.81_{\pm 0.00}$ & $0.00_{\pm 0.00}$ & $13.21_{\pm 0.34}$ & $1.27_{\pm 0.04}$ \\
            & ReAct & $31.43_{\pm 3.16}$ & $39.50_{\pm 0.62}$ & $\textbf{2.78}_{\pm 2.78}$ & $9.92_{\pm 0.62}$ & $\textbf{1.87}_{\pm 0.25}$ \\
            & \cellcolor{cyan!10}\textbf{\name (Ours)} 
            & \cellcolor{cyan!10}$\textbf{45.40}_{\pm 1.68}$ 
            & \cellcolor{cyan!10}$\textbf{51.17}_{\pm 3.38}$ 
            & \cellcolor{cyan!10}$\textbf{2.78}_{\pm 2.78}$ 
            & \cellcolor{cyan!10}$\textbf{19.01}_{\pm 0.62}$ 
            & \cellcolor{cyan!10}$1.42_{\pm 0.24}$ \\
        \midrule

        \multirow{6}{*}{Llama-3.1 8B} 
            & Zero-shot & $40.74_{\pm 0.67}$ & $48.29_{\pm 1.04}$ & $0.00_{\pm 0.00}$ & $15.53_{\pm 0.50}$ & $2.21_{\pm 0.14}$ \\
            & Chain of Thought & $37.72_{\pm 1.16}$ & $24.72_{\pm 2.50}$ & $0.00_{\pm 0.00}$ & $07.81_{\pm 0.87}$ & $2.22_{\pm 0.22}$ \\
            & Tree of Thought & $15.06_{\pm 0.69}$ & $15.52_{\pm 0.21}$ & $0.00_{\pm 0.00}$ & $08.64_{\pm 1.64}$ & $2.05_{\pm 0.15}$ \\
            & Prompt Explore & $43.07_{\pm 0.72}$ & $52.77_{\pm 2.08}$ & $0.00_{\pm 0.00}$ & $13.70_{\pm 0.59}$ & $1.60_{\pm 0.21}$ \\
            & ReAct & $22.92_{\pm 5.04}$ & $\textbf{60.38}_{\pm 1.95}$ & $0.00_{\pm 0.00}$ & $15.94_{\pm 2.34}$ & $\textbf{2.88}_{\pm 0.51}$ \\
            & \cellcolor{cyan!10}\textbf{\name (Ours)} 
            & \cellcolor{cyan!10}$\textbf{50.15}_{\pm 6.88}$ 
            & \cellcolor{cyan!10}$52.89_{\pm 3.51}$ 
            & \cellcolor{cyan!10}$\textbf{2.78}_{\pm 2.78}$ 
            & \cellcolor{cyan!10}$\textbf{16.85}_{\pm 0.58}$ 
            & \cellcolor{cyan!10}$2.17_{\pm 0.21}$ \\
        \midrule

        \multirow{6}{*}{Mistral Small 22B} 
            & Zero-shot & $62.25_{\pm 1.08}$ & $31.70_{\pm 2.08}$ & $0.00_{\pm 0.00}$ & $26.47_{\pm 0.57}$ & $1.35_{\pm 0.03}$ \\
            & Chain of Thought & $31.43_{\pm 2.22}$ & $25.16_{\pm 0.28}$ & $0.00_{\pm 0.00}$ & $09.63_{\pm 0.35}$ & $1.43_{\pm 0.13}$ \\
            & Tree of Thought & $32.54_{\pm 0.83}$ & $26.40_{\pm 1.71}$ & $0.00_{\pm 0.00}$ & $10.88_{\pm 0.85}$ & $1.27_{\pm 0.03}$ \\
            & Prompt Explore & $50.60_{\pm 0.69}$ & $42.46_{\pm 0.79}$ & $0.00_{\pm 0.00}$ & $21.48_{\pm 0.48}$ & $1.70_{\pm 0.05}$ \\
            & ReAct & $57.78_{\pm 5.47}$ & $23.54_{\pm 2.07}$ & $0.00_{\pm 0.00}$ & $26.70_{\pm 4.19}$ & $\textbf{3.37}_{\pm 0.02}$ \\
            & \cellcolor{cyan!10}\textbf{\name (Ours)} 
            & \cellcolor{cyan!10}$\textbf{74.28}_{\pm 2.99}$ 
            & \cellcolor{cyan!10}$\textbf{43.12}_{\pm 1.21}$ 
            & \cellcolor{cyan!10}$\textbf{2.78}_{\pm 2.78}$ 
            & \cellcolor{cyan!10}$\textbf{32.43}_{\pm 1.15}$ 
            & \cellcolor{cyan!10}$2.92_{\pm 0.66}$ \\
        \midrule
         \multirow{6}{*}{Llama-3.1 70B}
            & Zero-shot
            & $63.95_{\pm 2.00}$
            & $82.25_{\pm 0.75}$
            & $11.11_{\pm 5.56}$
            & $54.35_{\pm 1.29}$
            & $5.17_{\pm 0.10}$ \\
            & Chain of Thought
            & $61.29_{\pm 0.00}$
            & $80.23_{\pm 1.88}$
            & $5.56_{\pm 5.56}$
            & $45.23_{\pm 2.05}$
            & $\textbf{5.37}_{\pm 0.17}$ \\
            & Tree of Thought
            & $67.03_{\pm 4.55}$
            & $48.73_{\pm 3.70}$
            & $0.00_{\pm 0.00}$
            & $28.32_{\pm 3.47}$
            & $3.31_{\pm 0.44}$ \\
            & Prompt Explore
            & $61.95_{\pm 1.15}$
            & $83.87_{\pm 3.12}$
            & $0.00_{\pm 0.00}$
            & $52.51_{\pm 3.61}$
            & $\textbf{5.57}_{\pm 0.19}$ \\
            & ReAct
            & $57.03_{\pm 4.42}$
            & $70.60_{\pm 3.37}$
            & $\textbf{16.67}_{\pm 0.00}$
            & $44.79_{\pm 1.47}$
            & $4.72_{\pm 0.37}$ \\
            & \cellcolor{cyan!10}\textbf{\name (Ours)}
            & \cellcolor{cyan!10}$\textbf{71.12}_{\pm 2.24}$
            & \cellcolor{cyan!10}$\textbf{85.33}_{\pm 0.73}$
            & \cellcolor{cyan!10}$\textbf{16.67}_{\pm 0.00}$
            & \cellcolor{cyan!10}$\textbf{63.40}_{\pm 2.62}$
            & \cellcolor{cyan!10}$5.22_{\pm 0.09}$ \\
        \bottomrule
    \end{tabular}
    \vspace{-0.15cm}
    \caption{Performance of \name compared to baseline prompting strategies on TALES environments. Results report mean normalized scores across tasks in an environment$\pm$ standard error over 3 runs.}
    \label{tab:mainPerformance}
\end{table*}

\textbf{DORA Explorer Unveils New Information.} We measure environment coverage via the number of unique states (observations) visited per episode. Given a trajectory $\{o_t\}_{t=1}^T$, we define $\mathcal{S}_{\text{unique}} = \{o_t \mid t \in [T]\}$ and compute $|\mathcal{S}_{\text{unique}}|$. As shown in \autoref{fig:state_diversity}, DORA visits over $2\times$ more unique states than baselines in TextWorld and ScienceWorld, with a substantial gain in AlfWorld, suggesting more effective discovery of hidden objects and task-relevant signals.\\

\begin{table*}[t]
\centering
\small
\setlength{\tabcolsep}{4pt}
\renewcommand{\arraystretch}{1.2}
\begin{tabular}{l ccc ccc ccc}
\toprule
& \multicolumn{3}{c}{\textit{TextWorld}} & \multicolumn{3}{c}{\textit{ScienceWorld}} & \multicolumn{3}{c}{\textit{AlfWorld}} \\
\cmidrule(lr){2-4} \cmidrule(lr){5-7} \cmidrule(lr){8-10}
\textbf{Method} & \textbf{Loops} & \textbf{Loops} & \textbf{Rec.} & \textbf{Loops} & \textbf{Loops} & \textbf{Rec.} & \textbf{Loops} & \textbf{Loops} & \textbf{Rec.} \\
& \textbf{Enc.} & \textbf{Rec.} & \textbf{(\%)} & \textbf{Enc.} & \textbf{Rec.} & \textbf{(\%)} & \textbf{Enc.} & \textbf{Rec.} & \textbf{(\%)} \\
\midrule
Zero-shot       & 612           & 3           & 0.5           & 2551           & 9           & 0.35          & 811           & 0           & 0.0  \\
Prompt Explore  & 663           & 5           & 0.8           & 2321           & 10          & 0.43          & 721           & 1           & 0.1  \\
CoT             & 673           & 24          & 3.6           & 2621           & 6           & 0.23          & 846           & 1           & 0.1  \\
ToT             & 783           & 8           & 1.0           & 2801           & 2           & 0.07          & 857           & 0           & 0.0  \\
ReAct           & 656           & 14          & 2.13          & 2466           & 13          & 0.53          & 801           & 4           & 0.50 \\
DORA (Auto)     & \textbf{185}  & \textbf{36} & \textbf{19.5} & \textbf{1234}  & \textbf{45} & \textbf{3.6}  & \textbf{111}  & \textbf{12} & \textbf{10.8} \\
\bottomrule
\end{tabular}
\vspace{-0.15cm}
\caption{We report the number of loops encountered (\textit{o\textsubscript{t}}, \textit{a\textsubscript{t}} repeated) (cumulative across all tasks in an environment) for LLama 3.1 8B, loops recovered (whether the agent transitioned to a new state within the next 3 steps), and recovery rate (\%). Higher recovery indicates better ability to escape and avoid repetitive action cycles.}
\vspace{-0.45cm}
\label{tab:loop_recovery}
\end{table*}
\noindent\textbf{DORA Explore Enables Loop Recovery.} As shown in \autoref{tab:loop_recovery}, baseline methods frequently enter loops and rarely recover, with recovery rates typically below $1\%$. DORA reduces the number of loops encountered while achieving $5$--$20\times$ higher recovery rates. In TextWorld, DORA encounters only $185$ loops versus over $600$ for baseline methods, and recovers substantially more often.

We show in \autoref{tab:mainPerformance} that uncovering new information and preventing agents from becoming trapped in repetitive loops translates into consistent gains in downstream task success across most TALES environments. These gains closely follow how often the agent chooses to explore. As shown in \autoref{tab:explore_usage} and \autoref{fig:explore_frequency}, \textsc{Explore} is selected most frequently in AlfWorld, where feedback is uninformative (\textit{e.g.}, a frequent, generic response is \textit{`Nothing happens'}); DORA still makes measurable progress where baselines largely do not. Absolute performance, however, remains low in AlfWorld and Jericho across all methods, including DORA, \textit{i.e.}, improved exploration alone cannot overcome the limitations imposed by current model reasoning capabilities.

Overall, these results indicate that dynamic exploration with DORA is particularly effective for hidden-state discovery, uncertainty resolution, and recovery from early mistakes within limited horizons, leading to improved downstream task performance while incurring only a modest token overhead relative to baseline methods (\autoref{fig:tokenCount}).

\sectionvspace
\subsection{Ablations}
\label{sec:ablations}
\sectionvspace
We conduct a series of ablations on Llama 3.1 8B on key design factors in DORA and test whether its gains stem from structured exploration rather than generic stochasticity or diversity alone.\\
\vspace{-0.24cm}
\begin{table*}[ht]
\centering
\small
\setlength{\tabcolsep}{4pt}
\renewcommand{\arraystretch}{1.3}
\begin{adjustbox}{max width=\textwidth}
\begin{tabular}{l c c c c c c c c}
\toprule
\textbf{Model} 
& $\tau=0$ 
& $\tau=0.3$ 
& $\tau=0.7$ 
& $\tau=1$ 
& $\tau=1.5$ 
& $\tau=2$ 
& $\tau$-policy 
& \textbf{DORA} \\
\midrule
Llama-3.1 8B 
& 39.41 & 34.08 & 30.74 & 14.33 & 00.00 & 00.00 & 00.00 & \textbf{44.74} \\

Qwen-2.5 7B 
& \textbf{27.65} & 23.65 & 26.74 & 17.65 & 21.83 & 12.83 & 10.33 & \textbf{44.62} \\

Mistral Small 22B 
& 60.16 & 54.62 & 39.83 & 05.83 & 05.83 & 00.00 & 00.00 & \textbf{73.59} \\
\bottomrule
\end{tabular}
\end{adjustbox}
\vspace{-0.15cm}
\caption{
Effect of temperature ($\tau$) on performance in TALES TextWorld. 
$\tau$-policy denotes an exponentially decaying schedule from $\tau=2$ to $\tau=0$. 
DORA (Auto-Explore) consistently outperforms all temperature-based strategies, indicating that its gains arise from structured exploration rather than stochastic decoding.
}
\label{tab:temperature_ablation}
\end{table*}



\vspace{-0.10cm}
\noindent\textbf{Effect of $\alpha$.} Is the combined score actually necessary, or would either component alone suffice? The coefficient $\alpha$ controls the trade-off between the two scoring terms, weighting mean log-probability against log-probability variance. We tune $\alpha$ on MAB, where cumulative regret and suffix failure provide direct measures of exploration quality. As shown in Appendix~\ref{app:ablations} (Table~\ref{tab:alpha_beta_ablation}), intermediate value $\alpha=0.8$  achieves the best trade-off, yielding the lowest regret and highest average reward, and therefore use it throughout all experiments.\\

\noindent\textbf{Fixed $\lambda$-schedule vs.\ Auto-Explore.} We compare DORA's Auto-Explore mechanism with a fixed $\lambda$-schedule (Appendix~\ref{app:ablations}) (Table~\ref{tab:autoexplore_ablation}). While a fixed schedule performs well in MAB, where exploration is most critical early and becomes less necessary as learning progresses, it is substantially less effective in TALES, where exploration demand is triggered dynamically by the state (\textit{e.g.}\ loops) rather than following a predictable global trend.\\

\noindent\textbf{$\tau$-sampling vs.\ DORA.} We test whether generic stochastic decoding via temperature sampling can substitute for DORA's structured exploration on TALES's TextWorld environment (\autoref{tab:temperature_ablation}). Performance degrades as temperature increases, with greedy decoding performing best among temperature settings. Higher temperatures produce more invalid or incoherent actions. This indicates that DORA's gains do not come from randomness but from structured, sequence-level exploration.\\

\noindent\textbf{Does Score Matter, or Just Candidates?} To isolate whether DORA's gains come from generating diverse candidates or from the scoring-and-$\lambda$-sampling mechanism itself, we compare against a \emph{Random Sampling} variant that generates the same candidate set and explore/greedy decisions but samples uniformly from $\mathcal{C}$ instead of using Eq.~\ref{eq:lambdaProbabilities} (Appendix~\ref{app:ablations}) ({\autoref{tab:random_sampling}). Random Sampling performs substantially worse than both zero-shot and DORA despite exploring at a comparable rate, indicating that diverse candidate generation alone is insufficient: a meaningful share of DORA's gains comes from the scoring and $\lambda$-sampling mechanism ranking those candidates.

\begin{figure}[t]
\centering

\begin{tikzpicture}
\begin{axis}[
    xbar,
    width=\columnwidth,
    height=4.6cm,
    enlarge y limits={abs=0.35cm},
    xmin=0,
    xmax=400000,
    ylabel={},
    ytick=\empty,
    symbolic y coords={
        DORA,
        ReAct,
        ToT,
        CoT,
        Prompt Explore,
        Zero-shot
    },
    grid=both,
    major grid style={dashed, gray!55},
    minor grid style={gray!20},
    minor tick num=1,
    xtick={0,100000,200000,300000,400000},
    xticklabels={0,0.1M,0.2M,0.3M,0.4M},
    scaled x ticks=false,
    bar shift=0pt,
    bar width=9pt,
    legend image code/.code={
        \draw[#1]
        (0cm,-0.09cm) rectangle (0.28cm,0.09cm);
    },
    legend style={
        at={(0.5,1.05)},
        anchor=south,
        legend columns=3,
        draw=none,
        font=\footnotesize,
        column sep=0.12cm,
        row sep=1pt,
        cells={anchor=west}
    },
]

\addplot[
    fill=cyan!17,
    draw=black!80
]
coordinates {(203200.10,Zero-shot)};
\addlegendentry{Zero-shot}

\addplot[
    fill=orange!15,
    draw=black!80
]
coordinates {(193176.30,Prompt Explore)};
\addlegendentry{Prompt Explore}

\addplot[
    fill=teal!15,
    draw=black!80
]
coordinates {(283937.70,CoT)};
\addlegendentry{CoT}

\addplot[
    fill=blue!15,
    draw=black!80
]
coordinates {(297266.40,ToT)};
\addlegendentry{ToT}

\addplot[
    fill=red!15,
    draw=black!80
]
coordinates {(391571.00,ReAct)};
\addlegendentry{ReAct}

\addplot[
    fill=magenta!17,
    draw=black!80,
    postaction={
        pattern=north east lines,
        pattern color=black!70
    }
]
coordinates {(364664.10,DORA)};
\addlegendentry{DORA}

\end{axis}
\end{tikzpicture}

\vspace{-0.1cm}
\caption{
Average token usage per TextWorld task for Llama-3.1 8B.
DORA uses approximately $2\times$ more tokens than Prompt Explore while remaining less expensive than ReAct.
}
\label{fig:tokenCount}
\vspace{-0.55cm}

\end{figure}
\sectionvspace
\vspace{-0.15cm}
\section{Conclusion}
\vspace{-0.15cm}
\sectionvspace
One of the most important aspects of embodied agents is exploration: the ability to take unlikely paths and gather information from the environment. However, LLM agents struggle with this due to the disconnect between sampling in token space versus actions. To address this we introduce DORA Explorer, a general training-free LLM inference framework.
DORA goes beyond standard sampling and prompting strategies by exploring diverse action sequences in a structured manner. This enables the agent to reach more informative states, supporting rapid in-context adaptation. We show in our experiments across diverse environments that
DORA Explorer significantly outperforms sampling and prompting strategies and is able to generalize to different models and environments.
\newpage

\section*{Limitations}
DORA is a training-free method, and its performance is naturally bounded by the base model's own reasoning and instruction-following ability.
Because scoring relies on token-level log-probabilities, DORA is currently applicable to open-weight models and not to most closed-source, API-only LLMs. Finally, Auto-Explore's exploration triggering and $\lambda$ selection are currently learned through prompting rather than an explicit signal such as loop detection; pairing this with a more principled trigger or a learned value estimate is a promising direction for future work.

\section*{Reproducibility Statement}
We describe our implementation, hyperparameters, and environment setup in detail throughout the main paper and appendix to support reproduction of our results. Code is released alongside this submission via the project website.

\bibliography{custom}

\appendix

\section{Methodology Details}

\subsection{Min-Max Normalization in Scoring}
\label{subsec:minMaxNormalization}
The mean and variables of per-token log-probabilities are defined as

\begin{align}
    \mu(a) &= \frac{1}{N} \sum_{i\in a} \log P_\tau(i), \notag \\
    \sigma^2(a) &= \frac{1}{N} \sum_{i\in a} \big(\log P_\tau(i) - \mu(a)\big)^2,
\end{align}

with $\text{min-max}(x) := \frac{x - x_{\min}}{x_{\max} - x_{\min} + \epsilon}$, $\epsilon=10^{-8}$, computed over $\mathcal{C}$ at each step.

\subsection{HyperParameters}
\begin{table}[htbp]
\centering
\setlength{\tabcolsep}{6pt}
\renewcommand{\arraystretch}{1.2}
\resizebox{\columnwidth}{!}{%
\begin{tabular}{l c}
\toprule
\textbf{Hyperparameter} & \textbf{Value} \\
\midrule
Temperature of Decision Policy $\tau_d$ & 0.2 \\
Temperature of Candidate Action Policy $\tau_\mathcal{C}$ & 0.7 \\
Temperature of $\lambda$-sampling Policy $\tau_{\lambda}$ & 0.2 \\
Number of Candidate Answers $n_C$ & 20 \\
Scoring Hyperparameter $\alpha$ & 0.8 \\
Exponential Growth Constant $k$  & 5 \\
\bottomrule
\end{tabular}
}
\caption{Hyperparameters used for DORA Explorer.}
\label{tab:tales_hyperparams}
\end{table}
\section{Environment Details}
\subsection{Multi-Armed Bandits}
\label{app:mabsetup}

We study the hard instance from \citet{krishnamurthy2024can}, where a single optimal arm has mean $\mu^\star = 0.5 + \frac{\Delta}{2}$ and all other arms have mean $\mu = 0.5 - \frac{\Delta}{2}$, yielding a gap $\Delta = \mu^\star - \mu$. In our experiments, we fix $K = 5$, $\Delta = 0.2$, and horizon $T = 500$. We model the LLM as a decision-making agent that interacts with the bandit via natural language. At each step, the model is prompted with a description of the task (including the horizon $T$) and a concise summary of the interaction history. Actions correspond to selecting one of $K$ arms via a neutral button-pressing interface. The model is required to output a single action in a constrained format (\texttt{<Answer>...</Answer>}), specifying one arm. Outputs that do not follow the required format or fail to identify a valid arm are treated as invalid actions. Invalid actions yield no reward and incur a regret penalty equal to the suboptimality gap ($0.2$ in our setting). More details of the prompt design, including examples, are provided in Appendix~\ref{app:mab_prompt_design}. Classical bandit baselines (UCB, TS, Greedy, $\epsilon$-greedy) are averaged over $N = 1000$ runs, while LLM-based methods are averaged over $N = 20$ runs.

\subsection{TALES}
\label{app:talessetup}
Each TALES task is modeled as a partially observable Markov decision process (POMDP; \S\ref{sec:problemFormulation}), where the agent receives a textual observation along with interaction history and outputs a textual action at each step. Episodes terminate either upon task completion or after a fixed horizon of 100 steps.

We evaluate across all TALES environments, including TextWorld (10 tasks), TextWorldExpress (16 tasks), AlfWorld (12 tasks), ScienceWorld (30 tasks) and Jericho (54 tasks)
\section{Algorithms}
\subsection{Multi-Armed Bandit Algorithms}
\label{app:definitions}
\textbf{Upper Confidence Bound(UCB):} We use UCB ~\citep{auer2002finite} as an upper bound algorithm for MAB. UCB uses a balance of the exploitation and exploration factors. Score for each arm, the exploitation factor is equal to its empirical mean reward augmented by an exploration bonus. Specifically, the score for arm $a$ is defined as $\hat{\mu}_a + \sqrt{C / n_a}$, where $\hat{\mu}_a$ denotes the running average reward and $n_a$ is the number of times arm $a$ has been selected. At each step, the agent selects the arm with the highest score. To ensure sufficient initial coverage, each arm is pulled once before applying the UCB rule. In our implementation, we fix $C = 1$ as a simple and effective heuristic as also used by \citep{krishnamurthy2024can}
\textbf{Thompson sampling (TS):} We employ Thompson Sampling with independent Beta-Bernoulli models for each arm. The mean reward of each arm $a$ is treated as a latent parameter with a Beta prior, $\text{Beta}(\alpha_0, \beta_0)$, initialized uniformly with $\alpha_0 = \beta_0 = 1$. At each step, a sample is drawn from the posterior distribution $\text{Beta}(\alpha_a, \beta_a)$ for every arm, and the arm with the highest sampled value is selected. After observing a Bernoulli reward, the posterior is updated by incrementing $\alpha_a$ for a success and $\beta_a$ for a failure. This procedure naturally balances exploration and exploitation by sampling according to posterior uncertainty.
\textbf{Greedy:} After sampling each arm once, the greedy algorithm always selects the arm with the highest empirical mean reward. 
\textbf{$\epsilon$-greedy:} For $\epsilon$-greedy, at each step, the agent selects a random arm with probability $\epsilon$, and otherwise chooses the arm with the highest empirical mean reward. We initialize $\epsilon = 0.1$ and decay it multiplicatively by a factor of $0.99$ after each update to gradually transition from exploration to exploitation.
\section{Prompts}
\label{app:prompts}
\subsection{MAB Prompts}
\label{app:mab_prompt_design}
\begin{tcolorbox}[title={MAB System Prompt}, colback=gray!5, colframe=black!50, enhanced, breakable]
You are a bandit algorithm interacting with $K$ buttons labeled \{blue, green, red, yellow, purple\}. Each button is associated with a Bernoulli distribution with a fixed but unknown mean, which may differ across buttons.

At each time step, pressing a button yields a reward sampled from its associated distribution. You are given a total of $T$ time steps, and your goal is to maximize the cumulative reward over these $T$ steps.

After each step, you will be provided with a summary of your past actions and observed rewards. Based on this history, you must select exactly one button for the next step.

Your response must strictly follow this format:
\texttt{<Answer>I will press COLOR button</Answer>}
where COLOR is one of \{blue, green, red, yellow, purple\}.
\end{tcolorbox}

\begin{tcolorbox}[title={MAB Summarized History Prompt}, colback=gray!5, colframe=black!50, enhanced, breakable]
So far, you have taken $t$ actions. Your past choices and rewards are summarized below:

\begin{itemize}
    \item Blue button: pressed $n_{\text{blue}}$ times, average reward $\bar{r}_{\text{blue}}$
    \item Green button: pressed $n_{\text{green}}$ times, average reward $\bar{r}_{\text{green}}$
    \item Red button: pressed $n_{\text{red}}$ times, average reward $\bar{r}_{\text{red}}$
    \item Yellow button: pressed $n_{\text{yellow}}$ times, average reward $\bar{r}_{\text{yellow}}$
    \item Purple button: pressed $n_{\text{purple}}$ times, average reward $\bar{r}_{\text{purple}}$
\end{itemize}

If a button has not been pressed, its count is $0$ and no average reward is shown.

Which button will you choose next?

Your response must strictly follow this format:  
\texttt{<Answer>I will press COLOR button</Answer>}

where COLOR is one of \{blue, green, red, yellow, purple\}.
\end{tcolorbox}
\subsection{TALES Prompts}
\label{app:tales_prompt}
\begin{tcolorbox}[title=Zero-Shot System Prompt, colback=gray!5, colframe=black!50,enhanced, breakable]
You are playing a text-based game. Your goal is to maximize the final score.

At each step, read the current observation and output a single concise action to interact with the environment (e.g., get lamp). Do not include explanations.

If you are uncertain or stuck, consider using the "help" command to discover available actions.
\end{tcolorbox}
\begin{tcolorbox}[title={Zero-Shot CoT Prompt}, colback=gray!5, colframe=black!50, enhanced, breakable]
\ttfamily
You are playing a text-based adventure game. Your goal is to finish it with the highest score.\\

Before acting, reason briefly covering:\\
\quad 1. What the current scene tells you (exits, objects, state).\\
\quad 2. What your immediate sub-goal is.\\
\quad 3. Which single command best advances that sub-goal.\\

After your reasoning, output exactly one line and nothing after it:\\
\quad ACTION: <single short command, e.g. get lamp>\\

Rules:\\
\quad - Put only the game command after ACTION: (a few words). No second sentences on that line.\\
\quad - Do not write ACTION: anywhere except this final line.\\
\quad - No backticks or quotes around the command.\\
When stuck, try using the `help` command to see what commands are available.
\end{tcolorbox}
\begin{tcolorbox}[title={Tree-of-Thought (ToT) Prompt ($k=3$, $b=1$)}, colback=gray!5, colframe=black!50, enhanced, breakable]
\ttfamily

\textbf{Tree of Thought Reasoning}\\

\textbf{Branch Generation (k=3)}\\
Consider the current game state and generate EXACTLY 3 distinct candidate actions.\\
Each branch must differ meaningfully --- not just rephrasings of the same idea.\\
Format:\\
\quad Branch A: <action>\\
\quad Branch B: <action>\\
\quad Branch C: <action>\\

\textbf{Branch Evaluation}\\
For each branch, reason through ONE step ahead: what is the likely outcome?\\
Score each on two dimensions (1--5 each):\\
\quad - Progress: Does it advance toward the goal or unlock new options?\\
\quad - Safety: Does it avoid irreversible mistakes or dead ends?\\
Format:\\
\quad Branch A: Progress=? Safety=? Total=? \textbar\ Reasoning: <one sentence>\\
\quad Branch B: Progress=? Safety=? Total=? \textbar\ Reasoning: <one sentence>\\
\quad Branch C: Progress=? Safety=? Total=? \textbar\ Reasoning: <one sentence>\\

\textbf{Pruning (b=1)}\\
Eliminate the two lower-scoring branches.\\
State which branch survives and why in one sentence.\\

\textbf{Final Action}\\
After pruning, output exactly ONE line in this format (this line must be last):\\
\quad ACTION: <single short game command to type, e.g. open fridge>\\

Rules:\\
\quad - Put only the command after ACTION: (no backticks, no quotes).\\
\quad - Do not write ACTION: anywhere except this final line.\\
\quad - When stuck, try using the `help` command to see what commands are available.
\end{tcolorbox}
\begin{tcolorbox}[title=ReAct System Prompt, colback=gray!5, colframe=black!50,enhanced, breakable]
You are playing a text-based game and your goal is to finish it with the highest score.

At each step you receive an observation from the game. You must respond with a Thought and an Action, in that order.

Format your response exactly as:

Thought: brief reasoning about the situation and what to do  

Action: single short game command, e.g. get lamp

Rules:

- Always output exactly one Thought line followed by exactly one Action line.

- The Action must be a bare game command (no backticks, no quotes).

- When stuck, try using the "help" command to see what commands are available.
\end{tcolorbox}
\begin{tcolorbox}[title={Prompt Explore System Prompt}, colback=gray!5, colframe=black!50,enhanced, breakable]
You are playing a text-based game. Your goal is to maximize the final score.

Upon reading the current observation, output a single short phrase to interact with the game, e.g. get lamp, without backticks.

If you are stuck in an observation-action loop, or are uncertain about the next action, choose a more exploratory command and try different action phrases.

If you are stuck, consider using the "help" command to see what actions are available.
\end{tcolorbox}

\subsection{Explore prompts}
\begin{tcolorbox}[title={Mode Decision Prompt}, colback=gray!5, colframe=black!50, enhanced, breakable]
Decide whether to act GREEDY or EXPLORE.

You go GREEDY if you know your goal and how to achieve it, and you are confident about your next action.
Choose EXPLORE only if you are stuck in an observation-action loop, or are uncertain about your next action.

Return ONLY strict JSON on a single line with one of these shapes:
\texttt{\{"mode":"GREEDY"\}}
\texttt{\{"mode":"EXPLORE"\}}

Rules:
- mode must be exactly "GREEDY" or "EXPLORE".
- Do not include markdown, code fences, or explanation.
\end{tcolorbox}
\begin{tcolorbox}[title={Candidate Generation Prompt}, colback=gray!5, colframe=black!50,enhanced, breakable]
Based on the current game state, list \texttt{\{n\}} possible actions you could take.

Write one action per line as a short command phrase.

Do not number the lines. Do not include any explanation.
\end{tcolorbox}
\begin{tcolorbox}[title={Lambda Decision Prompt}, colback=gray!5, colframe=black!50, enhanced, breakable]
You chose to EXPLORE. Now select the exploration parameter $\lambda$.

$\lambda \in [\lambda_{\min}, \lambda_{\max}]$, where lower values correspond to higher exploration (flatter sampling), and higher values correspond to lower exploration (sharper sampling).

Return ONLY strict JSON on a single line with the following format:
\texttt{\{"lambda":<float>\}}

Rules:
- $\lambda$ must lie within the allowed range $[\lambda_{\min}, \lambda_{\max}]$.
- Example: \texttt{\{"lambda":0.5\}}
- Do not include any additional text, markdown, or explanation.
\end{tcolorbox}

\section{Additional TALES Findings}
\label{app:tales_results}
\textbf{How Often Does DORA Explore?} To assess whether DORA's gains stem from adaptive exploration rather than incidental diversity, we track how often Auto-Explore selects \textsc{Explore} across all TALES environments (\autoref{tab:explore_usage}; per-episode dynamics in \autoref{fig:explore_frequency}). Explore frequency varies substantially by environment and model, and is consistently high in AlfWorld and Jericho, the suite's two weakest-performing environments across all three models, and lower in environments closer to saturation, such as TextWorldExpress. When exploring, sampled $\lambda$ values have a median of $2.5$ (range $[0,10]$), indicating the agent favors flatter, more exploratory sampling once it has decided to explore, rather than sampling $\lambda$ near-greedily. This adaptive, state-dependent behavior is consistent with the loop-recovery gains in \autoref{tab:loop_recovery}: Auto-Explore is responding to signals of uncertainty rather than exploring at a fixed, environment-independent rate.
\begin{table*}[t]
\centering
\begin{tabular}{lccccc}
\toprule
\textbf{Model} & \textbf{TW} & \textbf{TW Exp.} & \textbf{AlfW} & \textbf{SciW} & \textbf{Jericho} \\
\midrule
Llama-3.1 8B  & 4.00\%  & 6.73\% & 10.94\% & 9.13\%  & 15.55\% \\
Qwen2.5 7B    & 42.43\% & 4.81\% & 67.75\% & 39.55\% & 12.87\% \\
Mistral 22B   & 17.40\% & 2.81\% & 27.81\% & 27.38\% & 17.63\% \\
Llama-3.1 70B & 21.22\% & 5.53\% & 70.19\% & 8.23\%  & 26.78\% \\
\bottomrule
\end{tabular}
\caption{Percentage of steps at which Auto-Explore selects \textsc{Explore}, by model and TALES environment.}
\label{tab:explore_usage}
\end{table*}

\begin{figure}[t]
    \centering
    \begin{tikzpicture}
\begin{axis}[
    width=1.05\columnwidth,
    height=0.72\columnwidth,
    title={
        Qwen2.5 --- \textsc{Explore} Frequency\\
        across TALES Environments
    },
    title style={
        align=center,
        font=\normalsize
    },
    xlabel={Episode Progress},
    ylabel={$P(\textsc{Explore})$},
    label style={font=\normalsize},
    tick label style={font=\small},
    xmin=0.8,
    xmax=5.2,
    ymin=0,
    ymax=1,
    xtick={1,2,3,4,5},
    xticklabels={
        {0--20},
        {20--40},
        {40--60},
        {60--80},
        {80--100}
    },
    ytick={0,0.2,0.4,0.6,0.8,1.0},
    grid=major,
    grid style={gray!25},
    axis line style={black},
    tick align=outside,
    line width=1.4pt,
    mark size=2.8pt,
    legend columns=2,
    legend cell align=left,
    legend style={
        at={(0.02,0.98)},
        anchor=north west,
        draw=none,
        fill=none,
        font=\scriptsize,
        column sep=6pt,
        row sep=1pt,
        cells={anchor=west}
    }
]

\addplot[
    color=plotblue,
    mark=*,
    mark options={fill=plotblue}
]
coordinates {
    (1,0.30)
    (2,0.365)
    (3,0.395)
    (4,0.365)
    (5,0.45)
};
\addlegendentry{TextWorld}

\addplot[
    color=plotorange,
    mark=*,
    mark options={fill=plotorange}
]
coordinates {
    (1,0.19)
    (2,0.12)
    (3,0.118)
    (4,0.128)
    (5,0.162)
};
\addlegendentry{TW Express}

\addplot[
    color=plotgreen,
    mark=*,
    mark options={fill=plotgreen}
]
coordinates {
    (1,0.375)
    (2,0.592)
    (3,0.705)
    (4,0.758)
    (5,0.77)
};
\addlegendentry{AlfWorld}

\addplot[
    color=plotred,
    mark=*,
    mark options={fill=plotred}
]
coordinates {
    (1,0.32)
    (2,0.333)
    (3,0.334)
    (4,0.335)
    (5,0.315)
};
\addlegendentry{ScienceWorld}

\addplot[
    color=plotpurple,
    mark=*,
    mark options={fill=plotpurple}
]
coordinates {
    (1,0.11)
    (2,0.075)
    (3,0.083)
    (4,0.078)
    (5,0.082)
};
\addlegendentry{Jericho}
\label{fig:explore_pct}
\end{axis}
\end{tikzpicture}

 
    \caption{
    Qwen2.5 \textsc{Explore} frequency over episode progress across TALES environments.
    }
    \label{fig:explore_frequency}
\end{figure}

\section{Additional MAB Results}
\label{app:mab_results}
\begin{table*}[!b]
\vspace{-0.1cm}
\centering
\setlength{\tabcolsep}{4pt}
\renewcommand{\arraystretch}{1.6}
\begin{adjustbox}{max width=\textwidth}
\begin{tabular}{l*{2}{c}|*{4}{c}|c}
\toprule
\multirow{2}{*}{\diagbox[width=8em,height=3em]{\textbf{Metric}}{\textbf{Method}}}
& \multicolumn{2}{c|}{\textit{Reference (Upper Bound)}}
& \multicolumn{4}{c|}{Baselines}
& DORA \\

\cmidrule(lr){2-3}
\cmidrule(lr){4-7}
\cmidrule(lr){8-8}

& UCB
& TS
& Greedy
& $\varepsilon$-Greedy
& $\tau=0$
& $\tau_\text{policy}$
& $\lambda_\text{exp}$ \\
\midrule

Mean Avg Reward
& \cellcolor{refGray}0.531
& \cellcolor{refGray}0.512
& \cellcolor{greenMed}\textbf{\textcolor{white}{0.499}}
& \cellcolor{greenLight}\textbf{0.489}
& \cellcolor{greenPale}\textbf{0.410}
& \cellcolor{greenPale}\textbf{0.410}
& \cellcolor{greenDark}\textbf{\textcolor{white}{0.519}}
\\

SuffFailFreq($T$/2)
& \cellcolor{refGray}0.017
& \cellcolor{refGray}0.002
& \cellcolor{greenMed}\textbf{\textcolor{white}{0.466}}
& \cellcolor{greenLight}\textbf{0.290}
& \cellcolor{greenPale}\textbf{0.900}
& \cellcolor{greenPale}\textbf{0.900}
& \cellcolor{greenDark}\textbf{\textcolor{white}{0.000}}
\\

Best arm Frac
& \cellcolor{refGray}0.658
& \cellcolor{refGray}0.560
& \cellcolor{greenMed}\textbf{\textcolor{white}{0.494}}
& \cellcolor{greenLight}\textbf{0.441}
& \cellcolor{greenPale}\textbf{0.100}
& \cellcolor{greenPale}\textbf{0.100}
& \cellcolor{greenDark}\textbf{\textcolor{white}{0.634}}
\\

Cum Regret
& \cellcolor{refGray}13.690
& \cellcolor{refGray}17.605
& \cellcolor{greenMed}\textbf{\textcolor{white}{20.254}}
& \cellcolor{greenLight}\textbf{22.363}
& \cellcolor{greenPale}\textbf{36.000}
& \cellcolor{greenPale}\textbf{36.000}
& \cellcolor{greenDark}\textbf{\textcolor{white}{14.660}}
\\

\bottomrule
\end{tabular}
\end{adjustbox}

\vspace{-0.25cm}
\caption{
Performance on the hard MAB instance using \texttt{Qwen2.5-7B Instruct}. \textit{Reference} columns (gray) report classical algorithms with theoretical guarantees. \textbf{DORA} consistently outperforms all baseline methods (green scale, darkest = best).
}
\vspace{-0.55cm}
\label{tab:mab_qwen}
\end{table*}

We report additional MAB results complementing \autoref{tab:mab} (Llama-3.1 8B, $T=500$) in the main text: Qwen2.5-7B on the same instance at $T=200$ (\autoref{tab:mab_qwen}), Llama-3.1 8B at $T=200$ (\autoref{tab:mab_200}), and a harder $9$-arm instance at $T=500$ (\autoref{tab:mab9arm}). Together, these confirm the main-text trends hold across models, horizons, and arm configurations.

On Qwen2.5-7B, DORA $\lambda_\text{exp}$ leads all LLM-based methods (mean reward 0.519, zero \textit{Suffix Failure Frequency}) and edges out TS on reward and best-arm fraction (0.634 vs.\ 0.560), while remaining below UCB.

On Llama-3.1 8B at $T=200$ (\autoref{tab:mab_200}), DORA again achieves the lowest regret and highest best-arm fraction among LLM-based methods, edging out TS, unlike at $T=500$ in the main text, where TS's guarantee grows stronger with more samples and both TS and UCB surpass DORA.

On the harder $9$-arm instance (\autoref{tab:mab9arm}), $\tau=0$ suffers most (\textit{Suffix Failure Frequency} 0.950, regret 131.250); DORA cuts failure frequency to 0.050 and roughly halves regret (45.900), while matching Greedy's reward (0.510), without any configuration-specific retuning.

\begin{table*}[t]
\vspace{-0.1cm}
\centering
\setlength{\tabcolsep}{4pt}
\renewcommand{\arraystretch}{1.6}
\begin{adjustbox}{max width=\textwidth}
\begin{tabular}{l*{2}{c}|*{7}{c}c}
\toprule
\multirow{2}{*}{\diagbox[width=8em, height=3em]{\textbf{Metric}}{\textbf{Method}}}
& \multicolumn{2}{c|}{\textit{Reference (Upper Bound)}}
& \multicolumn{7}{c}{Baselines}
& \multirow{2}{*}{\begin{tabular}{c}
\textbf{DORA}\\
\textbf{$\lambda_\text{exp}$}
\end{tabular}}
\\
\cmidrule(lr){2-3}
\cmidrule(lr){4-10}

& UCB
& TS
& Greedy
& $\varepsilon$-Greedy
& $\tau=0$
& $\tau=0.7$
& $\tau=1$
& $\tau=1.5$
& $\tau_\text{exp}$
& \\
\midrule

Mean Avg Reward
& \cellcolor{refGray}0.531
& \cellcolor{refGray}0.512
& \cellcolor{greenMed}\textbf{\textcolor{white}{0.499}}
& \cellcolor{greenLight}\textbf{0.489}
& \cellcolor{greenPale}\textbf{0.410}
& \cellcolor{greenPale}\textbf{0.410}
& \cellcolor{greenPale!50}\textbf{0.401}
& \cellcolor{greenLight!70}\textbf{0.440}
& \cellcolor{greenLight!50}\textbf{0.444}
& \cellcolor{greenDark}\textbf{\textcolor{white}{0.521}}
\\

SuffFailFreq($T$/2)
& \cellcolor{refGray}0.017
& \cellcolor{refGray}0.002
& \cellcolor{greenMed}\textbf{\textcolor{white}{0.466}}
& \cellcolor{greenLight}\textbf{0.290}
& \cellcolor{greenPale}\textbf{0.900}
& \cellcolor{greenPale}\textbf{0.900}
& \cellcolor{greenPale!50}\textbf{0.950}
& \cellcolor{greenLight!70}\textbf{0.250}
& \cellcolor{greenLight!50}\textbf{0.100}
& \cellcolor{greenDark}\textbf{\textcolor{white}{0.000}}
\\

Best arm Frac
& \cellcolor{refGray}0.658
& \cellcolor{refGray}0.560
& \cellcolor{greenMed}\textbf{\textcolor{white}{0.494}}
& \cellcolor{greenLight}\textbf{0.441}
& \cellcolor{greenPale}\textbf{0.100}
& \cellcolor{greenPale}\textbf{0.100}
& \cellcolor{greenPale!50}\textbf{0.040}
& \cellcolor{greenLight!70}\textbf{0.301}
& \cellcolor{greenLight!50}\textbf{0.505}
& \cellcolor{greenDark}\textbf{\textcolor{white}{0.618}}
\\

Cum Regret
& \cellcolor{refGray}13.690
& \cellcolor{refGray}17.605
& \cellcolor{greenMed}\textbf{\textcolor{white}{20.254}}
& \cellcolor{greenLight}\textbf{22.363}
& \cellcolor{greenPale}\textbf{36.000}
& \cellcolor{greenPale}\textbf{36.000}
& \cellcolor{greenPale!50}\textbf{38.390}
& \cellcolor{greenLight!70}\textbf{28.450}
& \cellcolor{greenLight!50}\textbf{21.690}
& \cellcolor{greenDark}\textbf{\textcolor{white}{15.260}}
\\

\bottomrule
\end{tabular}
\end{adjustbox}

\vspace{-0.25cm}
\caption{
Performance on the hard MAB instance using \texttt{Llama-3.1 8B} ($T=200$). Gray columns denote reference algorithms; darker green indicates better performance.
}
\vspace{-0.55cm}
\label{tab:mab_200}
\end{table*}

\vspace{0.3cm}
\begin{table*}[t]
\centering
\setlength{\tabcolsep}{4pt}
\renewcommand{\arraystretch}{1.6}
\begin{adjustbox}{max width=\textwidth}
\begin{tabular}{l*{2}{c}|*{3}{c}c}
\toprule
\multirow{2}{*}{\diagbox[width=8em,height=3em]{\textbf{Metric}}{\textbf{Method}}}
& \multicolumn{2}{c|}{\textit{Reference (Upper Bound)}}
& \multicolumn{3}{c}{Baselines}
& \multirow{2}{*}{
\begin{tabular}{c}
\textbf{DORA}\\
\textbf{$\lambda_{\text{exp}}$}
\end{tabular}}
\\

\cmidrule(lr){2-3}
\cmidrule(lr){4-6}

& UCB
& TS
& Greedy
& $\varepsilon$-Greedy
& $\tau=0$
& \\
\midrule

Mean Avg Reward
& \cellcolor{refGray}0.540
& \cellcolor{refGray}0.510
& \cellcolor{greenDark}\textbf{\textcolor{white}{0.510}}
& \cellcolor{greenLight}\textbf{0.490}
& \cellcolor{greenPale}\textbf{0.335}
& \cellcolor{greenDark}\textbf{\textcolor{white}{0.510}}
\\

SuffFailFreq($T$/2)
& \cellcolor{refGray}0.030
& \cellcolor{refGray}0.000
& \cellcolor{greenMed}\textbf{\textcolor{white}{0.520}}
& \cellcolor{greenLight}\textbf{0.420}
& \cellcolor{greenPale}\textbf{0.950}
& \cellcolor{greenDark}\textbf{\textcolor{white}{0.050}}
\\

Best arm Frac
& \cellcolor{refGray}0.650
& \cellcolor{refGray}0.550
& \cellcolor{greenMed}\textbf{\textcolor{white}{0.450}}
& \cellcolor{greenLight}\textbf{0.380}
& \cellcolor{greenPale}\textbf{0.050}
& \cellcolor{greenDark}\textbf{\textcolor{white}{0.502}}
\\

Cum Regret
& \cellcolor{refGray}30.780
& \cellcolor{refGray}43.320
& \cellcolor{greenMed}\textbf{\textcolor{white}{46.710}}
& \cellcolor{greenLight}\textbf{56.970}
& \cellcolor{greenPale}\textbf{131.250}
& \cellcolor{greenDark}\textbf{\textcolor{white}{45.900}}
\\

\bottomrule
\end{tabular}
\end{adjustbox}
\caption{
Performance on a $9$-arm Bernoulli MAB instance using
\texttt{Llama-3.1 8B} with horizon $T=500$. The arm reward means are
$\{0.10, 0.20, 0.30, 0.35, 0.40, 0.40, 0.45, 0.50, 0.60\}$.
\textit{Reference} columns (gray) report classical algorithms with theoretical guarantees.
\textbf{DORA} outperforms all LLM-based baselines (green scale, darkest = best).
}
\vspace{-0.55cm}
\label{tab:mab9arm}
\end{table*}
\FloatBarrier

\raggedbottom
\newpage
\clearpage
\begin{table*}[t]
\centering
\setlength{\tabcolsep}{6pt}
\renewcommand{\arraystretch}{1.4}
\begin{tabular}{l c c c c c c}
\toprule
\textbf{Metric}
& $(1.0, 0.0)$
& $(0.8, 0.2)$
& $(0.6, 0.4)$
& $(0.4, 0.6)$
& $(0.2, 0.8)$
& $(0.0, 1.0)$ \\
\midrule

Mean Avg Reward
& \cellcolor{greenLight!60}\textbf{0.496}
& \cellcolor{greenDark}\textbf{\textcolor{white}{0.521}}
& \cellcolor{greenMed}\textbf{0.509}
& \cellcolor{greenLight}\textbf{0.506}
& \cellcolor{greenLight!60}\textbf{0.500}
& \cellcolor{greenMed}\textbf{0.511} \\

Suff. Fail Freq ($T/2$)
& \cellcolor{greenLight!60}\textbf{0.05}
& \cellcolor{greenDark}\textbf{\textcolor{white}{0.00}}
& \cellcolor{greenLight!60}\textbf{0.05}
& \cellcolor{greenLight!60}\textbf{0.05}
& \cellcolor{greenLight!60}\textbf{0.05}
& \cellcolor{greenDark}\textbf{\textcolor{white}{0.00}} \\

Best Arm Frac
& \cellcolor{greenLight!60}\textbf{0.516}
& \cellcolor{greenDark}\textbf{\textcolor{white}{0.618}}
& \cellcolor{greenMed}\textbf{0.584}
& \cellcolor{greenLight}\textbf{0.540}
& \cellcolor{greenLight!60}\textbf{0.528}
& \cellcolor{greenLight}\textbf{0.577} \\

Cum. Regret
& \cellcolor{greenPale!80}\textbf{19.34}
& \cellcolor{greenDark}\textbf{\textcolor{white}{15.26}}
& \cellcolor{greenMed}\textbf{16.63}
& \cellcolor{greenLight!60}\textbf{18.40}
& \cellcolor{greenLight}\textbf{18.90}
& \cellcolor{greenLight}\textbf{16.92} \\

\bottomrule
\end{tabular}
\caption{
Ablation over explore-score weights $(\alpha, 1-\alpha)$ on the hard MAB instance using \texttt{Llama-3.1 8B}. Darker green indicates better performance. The updated results continue to show that $(0.8,0.2)$ provides the best overall trade-off.
}
\label{tab:alpha_beta_ablation}
\end{table*}

\begin{table}[htpb]
\centering
\setlength{\tabcolsep}{6pt}
\renewcommand{\arraystretch}{1.3}
\resizebox{\columnwidth}{!}{%
\begin{tabular}{lcc}
\toprule
\textbf{Model} &
\makecell{\textbf{DORA}\\\textbf{($\lambda$-schedule)}} &
\makecell{\textbf{DORA}\\\textbf{(Auto-Explore)}} \\
\midrule
Qwen-2.5 7B      & 5.83  & \textbf{42.95} \\
Llama-3.1 8B     & 21.00 & \textbf{41.89} \\
Mistral Small 22B & 9.17  & \textbf{69.57} \\
\bottomrule
\end{tabular}
}

\caption{
Comparison of DORA's fixed $\lambda$-schedule and Auto-Explore variants on the TALES TextWorld environment. Results are reported as normalized task scores from a single evaluation run.
}
\label{tab:autoexplore_ablation}
\end{table}
\begin{table}[h]
\centering
\begin{minipage}{0.48\textwidth}
\centering
\setlength{\tabcolsep}{6pt}
\renewcommand{\arraystretch}{1.4}
\resizebox{\textwidth}{!}{%
\begin{tabular}{l c c c}
\toprule
\textbf{Metric} & \textbf{$k=4$} & \textbf{$k=5$} & \textbf{$k=6$} \\
\midrule

Mean Avg Reward
  & \cellcolor{greenDark}\textbf{\textcolor{white}{0.524}}
  & \cellcolor{greenMed}\textbf{0.521}
  & \cellcolor{greenLight}\textbf{0.487}
\\

SuffFailFreq($T$/2)
  & \cellcolor{greenLight}\textbf{0.05}
  & \cellcolor{greenDark}\textbf{\textcolor{white}{0.00}}
  & \cellcolor{greenDark}\textbf{\textcolor{white}{0.00}}
\\

Best arm Frac
  & \cellcolor{greenDark}\textbf{\textcolor{white}{0.732}}
  & \cellcolor{greenMed}\textbf{0.618}
  & \cellcolor{greenLight}\textbf{0.557}
\\

Cum Regret
  & \cellcolor{greenDark}\textbf{\textcolor{white}{13.80}}
  & \cellcolor{greenMed}\textbf{15.26}
  & \cellcolor{greenLight}\textbf{17.73}
\\

\bottomrule
\end{tabular}
}
\caption{
Effect of the growth factor $k$ on the $\lambda$-schedule (cf.~\S\ref{subsec:selectLambda}). Larger $k$ favors exploration, while smaller $k$ favors exploitation; $k=5$ provides the best balance for \texttt{Llama-3.1 8B}.
}
\label{tab:growth_factor}
\end{minipage}
\end{table}

\section{Ablations}
\label{app:ablations}
We provide additional ablations examining the main design choices in DORA:
the action-scoring function, the growth rate of the fixed $\lambda$-schedule,
the use of state-dependent Auto-Explore, and the importance of score-based
candidate selection.
\paragraph{Effect of the score weight $\alpha$.}
We vary $(\alpha,1-\alpha)$, which balances the mean and variance of token
log-probabilities, on MAB using Llama-3.1 8B.

The setting $(0.8,0.2)$ achieves the best overall trade-off, with the highest
average reward, zero suffix failures, and the lowest cumulative regret. We therefore use $\alpha=0.8$ throughout.

\paragraph{Fixed $\lambda$-schedule versus Auto-Explore.}
A fixed schedule assumes that exploration should occur primarily at the
beginning of an episode and gradually decrease over time. This assumption
is suitable for MAB, but may not hold in sequential environments where the
need to explore depends on the current observation, uncertainty, or whether
the agent has entered a loop. We therefore compare the fixed schedule with
Auto-Explore, which dynamically decides when to explore and selects
$\lambda$ based on the interaction context.

Auto-Explore substantially outperforms the fixed schedule across all three
models on TextWorld. The improvement is especially large for Qwen-2.5 7B
and Mistral Small 22B, showing that exploration in TALES cannot be captured
well by a predetermined global schedule. Instead, exploration must be
triggered intermittently according to the agent's current state.

\paragraph{Effect of the $\lambda$-schedule growth factor.}
The growth factor $k$ in Eq.~\ref{eq:lambdaExpSchedule} determines how quickly
the fixed schedule transitions from exploration to exploitation. Smaller
values increase $\lambda$ more gradually, whereas larger values preserve
exploratory sampling for a greater portion of the episode.

With $k=4$, the policy achieves high reward and selects the best arm
frequently, but still exhibits occasional suffix failures. Increasing the
growth factor to $k=6$ removes suffix failures but retains exploration for
too long, reducing reward and increasing regret. We use $k=5$ as a more
reliable balance between discovering the optimal arm and subsequently
exploiting it.

\begin{table*}[!t]
\centering
\setlength{\tabcolsep}{3pt}
\begin{tabular}{lccccc}
\toprule
\textbf{Method} & \textbf{TW} & \textbf{TW Exp.} & \textbf{AlfW} & \textbf{SciW} & \textbf{Jericho} \\
\midrule
Zero-shot           & 29.20 & 48.93 & 0.00 & 13.49 & 0.66 \\
Random Sampling     & 15.33 & 44.88 & 0.00 & 11.47 & 0.45 \\
DORA ($\lambda$-sampling) & \textbf{45.40} & \textbf{51.17} & \textbf{2.78} & \textbf{19.01} & \textbf{1.42} \\
\midrule
Random Sampling Explore Use (\%) & 49.90 & 3.81 & 67.83 & 41.00 & 54.23 \\
\bottomrule
\end{tabular}
\caption{Isolating the scoring-and-$\lambda$-sampling mechanism from candidate diversity alone, on Qwen2.5-7B. Random Sampling generates the same diverse candidates and explore/greedy decisions as DORA, but samples uniformly from $\mathcal{C}$ instead of via Eq.~\ref{eq:lambdaProbabilities}. Bottom row reports the mean \% of steps at which Random Sampling chose to explore.}
\label{tab:random_sampling}
\end{table*}

\paragraph{Does scoring matter beyond candidate diversity?}
To separate the contribution of diverse candidate generation from that of
DORA's scoring and sampling mechanism, we construct a Random Sampling
variant. This variant uses the same candidate generation procedure and the
same greedy-versus-explore decisions as DORA, but selects uniformly from the
candidate set rather than sampling according to the score-derived
$\lambda$-probabilities.

Random Sampling performs considerably worse than DORA despite having access
to the same diverse candidate set. Thus, generating diverse actions alone
is insufficient: DORA's sequence-level scoring function and
$\lambda$-controlled sampling are necessary for prioritizing exploratory
actions that remain plausible and useful for task progress.

\end{document}